\documentclass[11pt]{article}
\usepackage{hyperref}
\usepackage{acl2019}
\usepackage{times}
\usepackage{latexsym}
\usepackage{amsmath}
\usepackage{mathtools}
\usepackage{booktabs}

\usepackage{url}
\PassOptionsToPackage{hyphens}{url}
\usepackage[english]{babel}
\usepackage{paralist}
\usepackage{latexsym}
\usepackage{tablefootnote}

\usepackage{ucs}
\usepackage{fontenc}
\usepackage{dsfont}
\usepackage{textcomp}
\usepackage{amsfonts}
\usepackage{graphicx}
\usepackage{sidecap}
\usepackage{pgfplots}
\pgfplotsset{compat=1.10}
\usepackage{array}
\usepackage{multirow}
\usepackage{bm}
\usepackage{breqn}
\usepackage{comment}

\aclfinalcopy 

\newcommand{\metriclong}[0]{POS-Discrimination Index}
\newcommand{\metric}[0]{PDI}
\newcommand{\blstm}[0]{Bi-LSTM}

\usepackage{collcell}
\makeatletter
\newcolumntype{G}{>{\collectcell\@gobble}c<{\endcollectcell}@{}}
\makeatother

\DeclareMathOperator*{\metricfunc}{PDI}

\title{Character Eyes: Seeing Language through Character-Level Taggers}

\author{Yuval Pinter \\
  School of Interactive Computing \\
  Georgia Institute of Technology \\
  {uvp@gatech.edu} \\\And
  Marc Marone \thanks{\enspace Work done while at Georgia Institute of Technology.} \\
  Microsoft Research\\ \\
  {v-mamaro@microsoft.com} \\\And
  Jacob Eisenstein \\
  Facebook AI Research \\ \\
  {jacobeisenstein@fb.com} \\}

\date{}

\begin{document}
\maketitle
\begin{abstract}
  Character-level models have been used extensively in recent years in NLP tasks as both supplements and replacements for closed-vocabulary token-level word representations.
  In one popular architecture, character-level LSTMs are used to feed token representations into a sequence tagger predicting token-level annotations such as part-of-speech (POS) tags.
  
  In this work, we examine the behavior of POS taggers across languages from the perspective of individual hidden units within the character LSTM.
  We aggregate the behavior of these units into language-level metrics which quantify the challenges that taggers face on languages with different morphological properties, and identify links between synthesis and affixation preference and emergent behavior of the hidden tagger layer.
  In a comparative experiment, we show how modifying the balance between forward and backward hidden units affects model arrangement and performance in these types of languages.
\end{abstract}

\section{Introduction}
\label{sec:intro}

Subword vector representations are now a standard part of neural architectures for natural language processing~\citep[e.g.,][]{fasttext, elmo}.
In particular, character representations have been shown to handle out-of-vocabulary words in supervised tagging tasks~\cite{ling2015finding, lample2016neural}.
These advantages generalize across multiple languages, where morphological formation may differ greatly but the character composition of words remains a relatively reliable primitive \cite{plank2016multiling}.

While the advantages of character-level models are readily apparent, existing evaluation methods fail to explain the mechanism by which these models encode linguistic knowledge about morphology and orthography.
Different languages exhibit character-word correspondence in very different patterns, and yet the bi-directional LSTM appears to be, or is assumed to be, capable of capturing them all.
In large multilingual settings, it is not uncommon to tune hyperparameters on a handful of languages, and apply them to the rest~\cite[e.g.,][]{mimick}.

In this work, we challenge this implicit generalization.
We train character-based sequence taggers
on a large selection of languages exhibiting various strategies for word formation, and subject the resulting models to a novel analysis of the behavior of individual units in the character-level \blstm{} hidden layer.
This reveals differences in the ability of the \blstm{} architecture to identify parts-of-speech, based on typological properties:
hidden layers trained on agglutinative languages find more regularities on the character level than in fusional languages; languages that are suffix-heavy give a stronger signal to the backward-facing hidden units, and vice versa for prefix-heavy languages.
In short, character-level recurrent networks function differently depending on how each language expresses morphosyntactic properties in characters.

These empirical results motivate a novel \blstm{} architecture, in which the number of hidden units is unbalanced across the forward and backward directions.
We find empirical correspondence between the analytical findings above and performance of such unbalanced \blstm{} models, allowing us to translate the typological properties of a language into concrete recommendations for model selection.
\footnote{\url{https://github.com/ruyimarone/character-eyes}}

\section{Related Work}
\label{sec:related}

Several recent papers attempt to explain neural network performance by investigating hidden state activation patterns on auxiliary or downstream tasks.
On the word level, \newcite{linzen2016assessing} trained LSTM language models, evaluated their performance on grammatical agreement detection, and analyzed activation patterns within specific hidden units.
We build on this analysis strategy as we aggregate (character-) sequence activation patterns across all hidden units in a model into quantitative measures.

Substantial prior work exists on the character level as well~\cite{karpathy2015visualizing, vania2017characters, kementchedjhieva2018indicatements, gerz2018language}.
\newcite{smith2018investigation} examined the character component in multilingual parsing models empirically, comparing it to the contribution of POS embeddings and pre-trained embeddings.
\newcite{chaudhary2018adapting} leveraged cross-lingual character-level correspondence to train NER models for low-resource languages.
Most related to our work is \newcite{godin2018explaining}, who compared CNN and LSTM character models on a type-level prediction task on three languages, using the post-network softmax values to see which models identify useful character sequences.
Unlike their analysis, we examine a more applied token-level task (POS tagging), and focus on the hidden states within the LSTM model in order to analyze its raw view of word composition.

Our analysis assumes a characterization of unit roles, where each hidden unit is observed to have some specific function. 
Findings from \newcite{linzen2016assessing} and others suggest that a single hidden unit can learn to track complex syntactic rules.
\newcite{radford2017learning} find that a character-level language model can implicitly assign a single unit to track sentiment, without being directly supervised. \cite{kementchedjhieva2018indicatements} also examine individual units in a character model and find complex behavior by inspecting activation patterns by hand.
In contrast, our metrics are motivated by discovering these units automatically, and capturing unit-level contributions quantitatively.

\begin{table}
    \centering
    \small
    \begin{tabular}{lGcccc}
        \toprule
        Language & \% N & Affix$^\dagger$ & Morph & \multicolumn{2}{c}{POS Accuracy \%} \\
        & & & synth$^\ddagger$ & Dev & Test \\ \midrule
        Arabic & 33.3 & S & int & 96.11 & 95.93\\
        Bulgarian & 21.8 & S & fus & 97.91 & 97.80\\
        Coptic & 14.0 & p & agg & 92.54 & 92.51\\
        Danish & 18.6 & S & fus & 95.59 & 95.46\\
        Greek & 21.2 & S & fus & 96.13 & 96.46\\
        English & 17.0 & S & fus & 93.65 & 93.30\\
        Spanish & 18.0 & S & fus & 95.75 & 95.00\\
        Basque & 24.4 & = & agg & 92.99 & 92.43\\
        Persian & 37.0 & s & fus & 96.07 & 96.10\\
        Irish & 27.0 & = & fus & & 89.35\\
        Hebrew & 23.6 & s & int & 95.71 & 94.60\\
        Hindi & 22.1 & S & fus & 95.03 & 94.91\\
        Hungarian & 22.4 & S & agg & 94.14 & 92.00\\
        Indonesian & 22.2 & S & iso & 92.55 & 92.68\\
        Italian & 19.9 & S & fus & 96.82 & 96.95\\
        Latvian & 26.8 & s & fus & 94.70 & 93.09\\
        Russian & 27.4 & S & fus & 95.29 & 95.25\\
        Swedish & 24.2 & S & fus & 95.80 & 95.73\\
        Tamil & 29.4 & S & agg & 86.46 & 87.58\\
        Thai & 27.4 & $\emptyset{}$ & fus & 91.37 &\\
        Turkish & 27.0 & S & agg & 92.08 & 92.48\\
        Ukrainian & 23.4 & S & fus & 95.68 & 95.26\\
        Vietnamese & 32.0 & $\emptyset{}$ & iso & 88.51 & 86.58\\
        Chinese & 27.5 & S & iso & 93.05 & 93.11\\
        \bottomrule
    \end{tabular}
    \caption{Attributes and tagging accuracy by language (Irish and Thai do not have both dev and test sets).
    $^\dagger$Affixation: S/s is strongly/weakly suffixing; P/p is strongly/weakly prefixing; = is equally prefixing/suffixing; $\emptyset{}$ is little affixation.
    $^\ddagger$Morphological synthesis: agglutinative, fusional, introflexive, isolating.}
    \label{tab:pos}
    \vspace{-2ex}
\end{table}

\section{Tagging Task}
\label{sec:task}

We train a set of LSTM tagging models, following the setup of \newcite{ling2015finding}.
A word representation trained from a character-level LSTM submodule is fed into a word-level bidirectional LSTM, with each word's hidden state subsequently fed into a two-layer perceptron producing tag scores, which are then softmaxed to produce a tagging distribution.
For languages with additional morphosyntactic attribute tagging, we follow the architecture in \newcite{mimick} where the same word-level \blstm{} states are used to predict each attribute's value using its own perceptron+softmax scaffolding.
In order to produce character models which would be as informative as possible to our subsequent analysis, we do not include word-level embeddings, pre-trained or otherwise, in our setup.

\subsection{Language Selection}
\label{ssec:langs}

As our goal is to examine the relationship between character-level modeling and linguistic properties, we drove language selection based on two morphological properties deemed relevant to the architectural effects examined.
All 24 datasets were obtained from Universal Dependencies (UD) version 2.3 \cite{ud23}, and linguistic properties were found in the World Atlas of Language Structures \cite{wals-20, wals-26}. The selected languages and their properties are presented in \autoref{tab:pos}.
We note that eleven of the 24 languages selected are not Indo-European.

\paragraph{Affixation.} To evaluate the role of forward and backward units in a bidirectional model, we selected all languages available in UD which are not classified as either weakly or strongly suffixing in inflectional morphology (the vast majority of UD languages).
This includes a single prefixing language (Coptic), two equally suffixing and prefixing languages (Basque and Irish), and two languages with little affixation (Thai and Vietnamese).

\paragraph{Morphological Synthesis.} Linguistically functional features vary between being expressed as distinct tokens (isolating languages), detectable unique character substrings (agglutinative), fused together but still distinguishable from the stem (fusional), and non-linearly represented within the word form (introflexive). This property has previously been found to affect performance in character-level models~\cite{mimick, gerz2018language, chaudhary2018adapting}, and thus we select representatives of each group, including most available non-fusional languages.

\subsection{Technical Setup}
Most of our selected languages have only a single UD 2.3 treebank.
For languages with multiple treebanks we selected the largest, except in the cases of Spanish and Indonesian, where we selected the GSD treebanks.
The Irish IDT treebank has only a train and test split, so we used the test set for early stopping.
The Thai PUD treebank only provided a single dataset with 1000 instances, which we shuffled and partitioned into a 850/150 split.
Tokens were normalized to remove noisy data: tokens containing `http' were replaced with `URL' and tokens containing `@' were replaced with `EMAIL'. This was most relevant (293 replacements) for the English treebank, which contained many long URLs. 

\paragraph{Hyperparameters.}
For the initial bidirectional character-level LSTM, we used a total hidden state size of 128 (64 units in each direction).
The character embedding size is set to 256, initialized using the method of \newcite{glorot2010understanding}.
The word-level bidirectional LSTM has two layers and a hidden state size of 128, with 50\% dropout applied in the style of \newcite{NIPS2016_6241}.
Each attribute-prediction MLP has a single hidden layer that is the same size as the tagset size for that attribute, and includes a $\tanh$ nonlinearity.
Models were trained for up to 80 epochs, and we select the model with the highest POS tagging accuracy on the dev set.
Training used SGD with 0.9 momentum, and all models were implemented using DyNet 2.0 \cite{dynet}.

\subsection{Results}
\label{ssec:results}

In our initial setup, we represent words using a concatenation of the final states from a bidirectional character-level LSTM with 64 forward and backward hidden units each.
The results for POS tagging, presented in \autoref{tab:pos}, are on par with similar models \cite[for example]{plank2016multiling} despite not including a word-level type embedding component.
We attribute this success to our large character embedding size of 256, corroborating findings reported by \newcite{smith2018investigation}.

\section{Analysis}
\label{sec:analysis}

We next analyze the models trained on the tagging task in an attempt to see how their character-level hidden states encode different manifestations of linguistic information.
We suggest that individual hidden units in the character-level sequence model attune to track patterns in the words which would indicate their linguistic roles (POS and morphological properties), and so patterns in character-role regularity across typologically different languages would manifest themselves in an observable form at the individual unit activation level.
This motivates us to devise metrics which would characterize languages through aggregation of individual unit behaviour.

\subsection{Metrics}

For each language, we run the character-level BiLSTM from the trained tagger on POS-unambiguous word types occurring frequently in the training set, grouped into their parts of speech.\footnote{We used 8 as our frequency threshold, and define unambiguous forms as ones tagged at least 60\% of the time with a single POS.}
This filtering was done in order to focus on the more consistent generalizations found by the taggers during training, as our goal is to qualify properties of languages.\footnote{This consideration also motivated our choice of UD data, which is tokenized to separate syntactic fusion such as Hebrew and Arabic function words, or Spanish \textit{del}.}
On each word $w$, we observe each hidden unit $h_i$'s activation level (output) on each character $h_i^c$.
We obtain a \textbf{base measure} $b(w,i)$ based on the activation pattern.
For example, an \textit{average absolute} base measure is defined as the average of absolute value activations:
$$b_{\text{avg}|\cdot|}(w,i) = \frac{1}{|w|} \sum_{c=1}^{|w|} |h_i^c|.$$
The \textit{max absolute diff} base measure is defined as:
$$b_{\text{mad}}(w,i) = \max_{c=1}^{|w|-1} |h_i^{c+1}-h_i^c|.$$
\begin{figure}
    \centering
    \frame{\includegraphics[width=75mm]{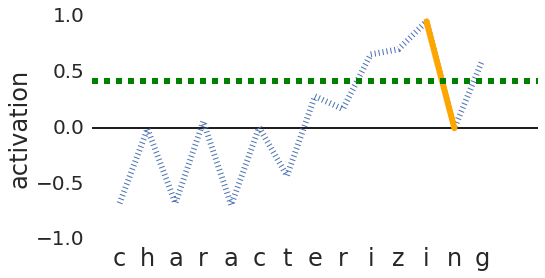}}
    \caption{Activations of the English model's unit $42$ (forward) on the word \textit{characterizing}. $b_{\text{avg}|\cdot|}$ is 0.42, and
    $b_{\text{mad}}$ is 0.96 (the drop from the second \textit{i} to \textit{n}).}
    \label{fig:ex}
\end{figure}

\autoref{fig:ex} demonstrates these two metrics for a sample (word, unit) pair, showing how the former captures the general level of activation the word caused on the unit, while the latter captures the local character pattern deemed most important by it.
We intentionally did not consider metrics based on the final activation values, the direct signals used by the later layers in the model, as these bear no insight into the effect of a word's composition on the learned model.

Next, we derive a language-level metric for each hidden unit, based on the principle of Mutual Information (MI).
The base metric's range ($[0,1)$ for $b_{\text{avg}|\cdot|}$, $[0,2)$ for $b_{\text{mad}}$) is divided into $B$ bins of equal size, and base activations from each word are summed across each of the $T$ POS tag categories\footnote{We omit the following  `character-simple' part-of-speech tags: INTJ, NUM, PROPN, PUNCT, SYM, X.}, then normalized to produce a joint probability distribution.
The mutual information is computed as:
$$    \sum_{t=1}^{T} \sum_{b=1}^{B} P(t, b)  \big[\ln P(t, b) - \ln P(t) - \ln P(b)\big], $$
and we call the resulting number the \metriclong{}, or \textbf{\metric{}}.
Intuitively, a higher \metric{} implies that the unit activates differently on words of different parts of speech, i.e. it is a better discriminator for the task.

At this point a language produces a set of $d_h$ PDI scores, one for each unit.
We sort them from high to low, and define two language-level metrics:
The \textbf{mass} is the sum of \metric{} values for all units,  
$\mathcal{M}(\mathcal{L}) := \sum_{i=1}^{d_h} \metricfunc (\mathcal{L}, i),$
intuitively meant to quantify the degree of success the model has in assigning hidden units to discriminate POS in this language.
The \textbf{head forwardness} is the proportion of forward-directional units before the point at which half of the mass accumulates (in a random setup, this number would tend to 0.5):
$$ \frac{\left|\left\{k:\sum_{i=1}^{k} \metricfunc (\mathcal{L}, i) \leq \frac{\mathcal{M}(\mathcal{L})}{2} \wedge h_k ~ \text{is forward}\right\}\right|}{\left|\left\{k:\sum_{i=1}^{k} \metricfunc (\mathcal{L}, i) \leq \frac{\mathcal{M}(\mathcal{L})}{2}\right\}\right|} $$
This metric aims to quantify the relative importance of forward and backward units in discriminating POS for $\mathcal{L}$.

\subsection{\metric{} Patterns}
\label{ssec:patt}

\begin{figure*}
    \centering
    \includegraphics[width=76mm]{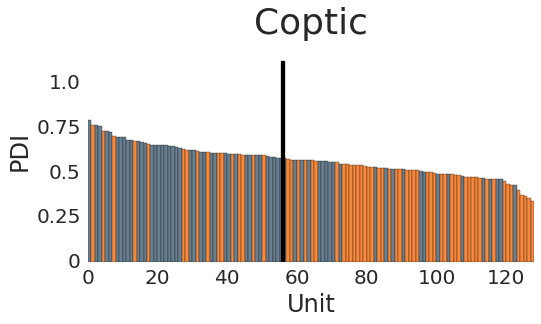} ~~~~
    \includegraphics[width=76mm]{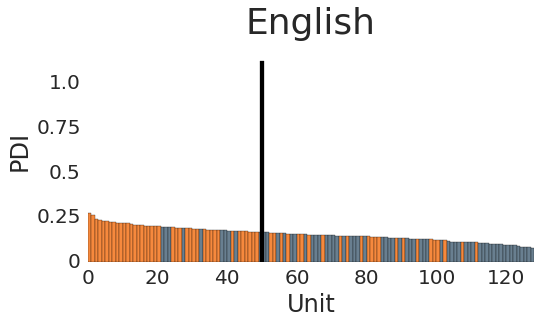}
    \caption{Distribution of \metric{} values ($b_{\text{avg}|\cdot|}$) across hidden units in Coptic and English, shown in ordered \metric{} values from largest to smallest, with blue (orange) bars indicating forward (backward) units.
    The black line demarcates the median point of mass accumulation.}
    \label{fig:pdis}
\end{figure*}

\begin{table}
    \centering
    \small
    \begin{tabular}{lccGc}\toprule
        Language & Mass & Mass & & \% of forward \\
        & & median & & units until \\
        & & index & & median \\ \midrule
        \textbf{Tamil} & 71.0 & 55 & 49.4 & 49.1 \\
        Irish & 62.0 & 56 & 48.6 & 42.9 \\
        \textbf{Coptic} & 58.1 & 56 & 52.9 & 71.4 \\
        \textbf{Hungarian} & 47.9 & 55 & 50.9 & 50.9 \\
        Greek & 31.2 & 55 & 49.1 & 45.5 \\
        \textbf{Turkish} & 30.1 & 54 & 51.5 & 57.4 \\
        Russian & 25.9 & 54 & 48.9 & 40.7 \\
        Thai & 25.9 & 55 & 50.3 & 47.3 \\
        Ukrainian & 25.0 & 54 & 47.7 & 37.0 \\
        Vietnamese & 24.2 & 55 & 48.3 & 36.4 \\
        Chinese & 23.8 & 47 & 47.2 & 42.6 \\
        Danish & 21.7 & 54 & 49.2 & 44.4 \\
        Swedish & 20.8 & 53 & 47.1 & 34.0 \\
        \textbf{Basque} & 20.6 & 51 & 53.6 & 64.7 \\
        Indonesian & 20.3 & 45 & 57.9 & 71.1 \\
        Latvian & 17.0 & 52 & 46.7 & 42.3 \\
        Spanish & 16.1 & 45 & 43.5 & 33.3 \\
        English & 16.0 & 50 & 43.0 & 20.0 \\
        Bulgarian & 15.6 & 52 & 48.0 & 46.2 \\
        Italian & 14.1 & 48 & 51.8 & 56.2 \\
        \textit{Arabic} & 12.6 & 46 & 52.7 & 58.7 \\
        \textit{Hebrew} & 11.4 & 51 & 55.4 & 74.5 \\
        Persian & 10.3 & 50 & 49.5 & 46.0 \\
        Hindi & 8.4 & 51 & 48.4 & 41.2 \\
        \bottomrule
    \end{tabular}
    \caption{\metric{} statistics for UD 2.3 models, $b_{\text{avg}|\cdot|}$ metric, sorted by the mass metric (sum of \metric{}s).
    Agglutinative languages in \textbf{bold}, introflexive in \textit{italics}.}
    \label{tab:pdipmi}
\end{table}

The \metric{} patterns on the $b_{\text{avg}|\cdot|}$ base measure with $B=16$ bins on all 24 languages are presented in \autoref{tab:pdipmi}.
We see that agglutinative languages, where we can expect a better discrimination signal to emerge from the consistently-formed morphemes, cluster mostly at the top of the \metric{} mass scale, suggesting more individual character-level units extract these signals successfully.
Introflexive languages, where character sequences seldom correspond to useful indications of POS or morphosyntactic attributes, cluster towards the bottom.

We present the full unit-level \metric{} value distributions for Coptic, a prefixing agglutinative language, and English, a suffixing fusional language, in \autoref{fig:pdis} (trends for $b_{\text{mad}}$ are similar).
Consistent with other agglutinative languages, Coptic's cumulative mass is very large ($\mathcal{M(\text{cop})}=58.1$), suggesting the predictive qualities of the sequence-based LSTM allows good discrimination from the character signal, as one might expect from an agglutinative language.
Conversely, $\mathcal{M(\text{eng})}=16$, demonstrating the difficulty presented by fusional languages.
The accumulation of 71\% forward (80\% backward) units in the head of the Coptic (English) value ranking suggests an interesting relationship between affixation and LSTM direction: LSTM units are likely to hone in on POS-indicative signals, which often occur as affixes, in the beginning of their run, causing activation values to rise (in absolute value) and stay large throughout the subsequent traversal of the stem.
Unfortunately, since no other prefixing languages are available in UD, we were not able to pursue this hypothesis further.

\subsection{Asymmetric Directionality}
\label{ssec:bal}

\begin{table}
    \centering
    \small
    \begin{tabular}{lrrrrr}\toprule
        Language & 128/0 & 96/32 & 64/64 & 32/96 & 0/128 \\
        Type & & & (base) \\ \midrule
        \multicolumn{6}{c}{Inflectional Affixation Categories} \\ \midrule
        S. suffix & \textbf{+0.22} & +0.07 & 94.50 & -0.06 & -0.02 \\
        W. suffix & +0.26 & \textbf{+0.12} & 95.46 & -0.07 & -0.01 \\
        Equal p/s & \textbf{+0.61} & +0.32 & 90.99 & -0.07 & +0.06 \\
        Little aff. & -0.06 & -0.21 & 89.59 & -0.16 & -0.22 \\
        W. prefix & \textbf{+0.52} & \textbf{+0.22} & 92.91 & \textbf{+0.40} & \textbf{+0.33} \\
        \midrule
        \multicolumn{6}{c}{Morphological Synthesis Categories} \\ \midrule
        Introflex. & \textbf{+0.17} & +0.05 & 95.87 & -0.06 & +0.01 \\
        Fusional & \textbf{+0.22} & +0.07 & 94.95 & +0.01 & +0.06 \\
        Agglutina. & \textbf{+0.59} & \textbf{+0.27} & 91.58 & -0.16 & -0.15 \\
        Isolating & -0.14 & -0.13 & 91.15 & -0.15 & -0.13 \\ \midrule
        Overall & \textbf{+0.25} & \textbf{+0.08} & 93.85 & -0.05 & -0.01 \\
        \bottomrule
    \end{tabular}
    \caption{Imbalanced models' mean POS accuracy on UD development data (differences between three averaged random runs in all models; \textbf{boldfaced} when significant at $p<0.05$ using a paired two-tailed \textit{t}-test).}
    \label{tab:abl}
\end{table}

Based on these observations, we conduct a directionality balance study, where we vary the number of hidden units in the forward and backwards dimensions.
In addition to the models analyzed above, which use 64 forward and 64 backward units (denoted hereafter 64/64), we trained models with imbalanced directionality (128/0, 96/32, 32/96, 0/128).
We test the hypothesis that imbalanced models affect languages differently based on their linguistic properties and statistical metrics.
We note that these settings do not maintain parameter set size: intra-direction transition operations are quadratic in that direction's hidden layer size, and so this adds a possible advantage in favor of direction-imbalanced models.

The results for this study are presented in \autoref{tab:abl} as averages for the language categories listed in \autoref{tab:pos} (the full, raw results are available in \autoref{tab:full_imb}).
\begin{table}
    \centering
    \small
    \begin{tabular}{lrrrrr} \toprule
        Language & 128/0 & 96/32 & 64/64 & 32/96 & 0/128 \\ \midrule
        Arabic & 96.29 & 96.08 & 96.06 & 96.09 & 96.16 \\
        Bulgarian & 97.95 & 97.86 & 97.84 & 97.74 & 97.71 \\
        Coptic & 93.10 & 92.80 & 92.58 & 92.98 & 92.91 \\
        Danish & 95.93 & 95.68 & 95.61 & 95.60 & 95.70 \\
        Greek & 96.19 & 96.07 & 96.01 & 96.00 & 95.93 \\
        English & 93.86 & 93.74 & 93.65 & 93.80 & 93.87 \\
        Spanish & 95.74 & 95.63 & 95.64 & 95.64 & 95.77 \\
        Basque & 93.52 & 93.13 & 92.89 & 92.59 & 92.90 \\
        Persian & 96.31 & 96.20 & 96.11 & 96.02 & 96.20 \\
        Irish & 89.54 & 89.35 & 88.95 & 89.11 & 89.07 \\
        Hebrew & 95.76 & 95.72 & 95.60 & 95.50 & 95.57 \\
        Hindi & 95.35 & 95.22 & 95.12 & 95.11 & 95.25 \\
        Hungarian & 94.25 & 94.29 & 94.20 & 93.97 & 94.00 \\
        Indonesian & 92.42 & 92.34 & 92.49 & 92.53 & 92.55 \\
        Italian & 97.00 & 96.78 & 96.87 & 96.88 & 97.01 \\
        Latvian & 95.10 & 94.84 & 94.69 & 94.58 & 94.61 \\
        Russian & 95.51 & 95.39 & 95.32 & 95.31 & 95.36 \\
        Swedish & 95.93 & 95.69 & 95.64 & 95.52 & 95.85 \\
        Tamil & 87.54 & 87.28 & 86.88 & 86.28 & 85.99 \\
        Thai & 91.52 & 91.27 & 91.38 & 91.47 & 91.32 \\
        Turkish & 93.14 & 92.45 & 92.06 & 92.03 & 92.09 \\
        Ukrainian & 95.72 & 95.76 & 95.63 & 95.68 & 95.66 \\
        Vietnamese & 87.98 & 87.92 & 88.23 & 87.83 & 87.85 \\
        Chinese & 93.01 & 93.17 & 93.12 & 93.03 & 93.04 \\ \bottomrule
    \end{tabular}
    \caption{Full scores for the directionality balance experiment, each point averaged over three random seed runs.}
    \label{tab:full_imb}
\end{table}


One trend which emerges is the preference of \textbf{agglutinative} languages for imbalanced models, whereas the other languages are little affected by this change.
This could be explained by the increase in inter-unit interaction in the larger direction of an imbalanced model -- contiguous character sequences consistently code reliable linguistic features in these languages.
A second finding is the slight bias of suffixing languages towards more forward units and of the prefixing language to more backward units, indicating that hidden LSTM units are better in detecting formations close to their final state.
Coupled with the findings regarding \metric{} mass distribution in the different directional units in \autoref{ssec:patt}, we suggest that a subtle relation exists between morphological information and model directionality: units which end their run on the affix are more important for detecting the POS signal, but it is more challenging for them to do so, and as a result more of them are necessary.
We also note the stability of isolating and little-affixing languages to directionality balance, possibly owing to the relatively small significance of contiguous character sequences in detecting word role.
Lastly, we point out that the compromise \textit{sesquidirectional} models 96/32 and 32/96 did not tend to stand out significantly on our tested language categories, suggesting there is no substantial middle-ground between the two popular techniques of unidirectional and bidirectional LSTMs.


\section{Conclusion}
While character-level \blstm{} models compute meaningful word representations across many languages, the way they do it depends on each language's typological properties.
These observations can guide model selection: for example, in agglutinative languages we observe a strong preference for a single direction of analysis, motivating the use of unidirectional character-level LSTMs for at least this type of language.

\section*{Acknowledgments}
We would like to thank Sebastian Mielke, anonymous NAACL reviewers, and the members of the Computational Linguistics lab at Georgia Tech for their valuable notes.
YP is a Bloomberg Data Science PhD Fellow. MM's work was funded by the Georgia Tech Undergraduate Research Opportunities Program.

\bibliography{char_eyes}
\bibliographystyle{acl_natbib}

\end{document}